\documentclass[manuscript,nonacm]{acmart}

\AtBeginDocument{%
  }

\usepackage{booktabs}
\usepackage{multirow}
\usepackage{amsmath}
\usepackage{xcolor}
\usepackage{tabularx}
\usepackage{array}
\usepackage[firstpage]{draftwatermark}  %
\SetWatermarkText{PREPRINT}
\SetWatermarkScale{0.8}
\SetWatermarkColor[gray]{0.92}    
\newcommand{\out}[1]{}

\copyrightyear{2027}
\acmYear{2027}
\setcopyright{acmlicensed}
\acmDOI{XXXXXXX.XXXXXXX}
\acmConference[CHI '27]{the CHI Conference on Human Factors
  in Computing Systems}{May 10--14, 2027}{Pittsburgh, PA, USA}
\acmISBN{978-1-4503-XXXX-X/2027/05}

\newif\ifmigrationnotes
\migrationnotestrue

\begin{document}

\title{Right Words, Wrong Moment: A Clinician-Grounded Analysis of Distress in 19,930 Conversations between Young People and ChatGPT}

\author{Marx Wang}
\affiliation{\department{Information School}\institution{University of Washington}\city{Seattle}\state{Washington}\country{United States}}
\email{marxwang@uw.edu}

\author{Ella Zhang}
\affiliation{\department{Information School}\institution{University of Washington}\city{Seattle}\state{Washington}\country{United States}}
\email{ezhang43@uw.edu}

\author{Cameron Tan}
\affiliation{\institution{University of Oxford}\city{Oxford}\country{United Kingdom}}
\email{cameron.tan@gtc.ox.ac.uk}

\author{Andrea Mock}
\affiliation{\institution{Stanford University}\city{Stanford}\state{California}\country{United States}}
\email{amock@stanford.edu}

\author{Songling Ngo}
\affiliation{\department{Information School}\institution{University of Washington}\city{Seattle}\state{Washington}\country{United States}}

\author{Zijing Wang}
\affiliation{\institution{University of Washington}\city{Seattle}\state{Washington}\country{United States}}

\author{Robert Wolfe}
\affiliation{\institution{University of Washington}\city{Seattle}\state{Washington}\country{United States}}

\author{Shirin Amouei}
\affiliation{\institution{University of Washington}\city{Seattle}\state{Washington}\country{United States}}
\email{shirinam@uw.edu}

\author{Rachel A. Hanebutt}
\affiliation{\department{Psychiatry, School of Medicine}\institution{Georgetown University}\city{Washington}\state{District of Columbia}\country{United States}}
\email{rachel.hanebutt@georgetown.edu}

\author{Desmond C. Ong}
\affiliation{\department{Department of Psychology}\institution{The University of Texas at Austin}\city{Austin}\state{Texas}\country{United States}}
\email{desmond.ong@utexas.edu}

\author{Caroline Figueroa}
\affiliation{\department{School of Medicine}\institution{Stanford University}\city{Palo Alto}\state{California}\country{United States}}
\email{cfiguer@stanford.edu}

\author{Katie Davis}
\affiliation{\institution{University of Washington}\city{Seattle}\state{Washington}\country{United States}}

\author{Anind K. Dey}
\affiliation{\department{Information School}\institution{University of Washington}\city{Seattle}\state{Washington}\country{United States}}

\author{Alexis Hiniker}
\affiliation{\department{Information School}\institution{University of Washington}\city{Seattle}\state{Washington}\country{United States}}

\renewcommand{\shortauthors}{Wang et al.}
\begin{abstract}
Young people increasingly turn to General-Purpose Conversational Agents (GPCAs), such as ChatGPT, in moments of distress. We examine young adults' (age 18--25) experiences using ChatGPT. 
We first collected 19,930 ChatGPT conversations and survey data from 158 young adults. We then selected five example conversations reflecting user distress. 
Finally, we asked ten clinicians to review those five conversations. 
We found distressed participants reported greater emotional engagement with ChatGPT and greater behavioral change from using it than their peers. When they turned to ChatGPT in moments of acute distress, ChatGPT was quick to give overly dramatic responses and excessive action-oriented suggestions. 
Clinicians endorsed ChatGPT's availability and much of its wording, but identified seven process failures, such as prematurely jumping to solutions. 
We translated clinicians' feedback into design guidelines following three stages: 1) asking about safety, 2) de-escalating intensity to restore emotional regulation, and 3) exploring concerns without agreeing with them.

\end{abstract}

\ccsdesc[500]{Human-centered computing~Empirical studies in HCI}

\keywords{Conversational agents, ChatGPT, psychological distress, young adults, self-disclosure}

\maketitle

\begin{figure}[t]
\centering
\includegraphics[width=0.9\linewidth]{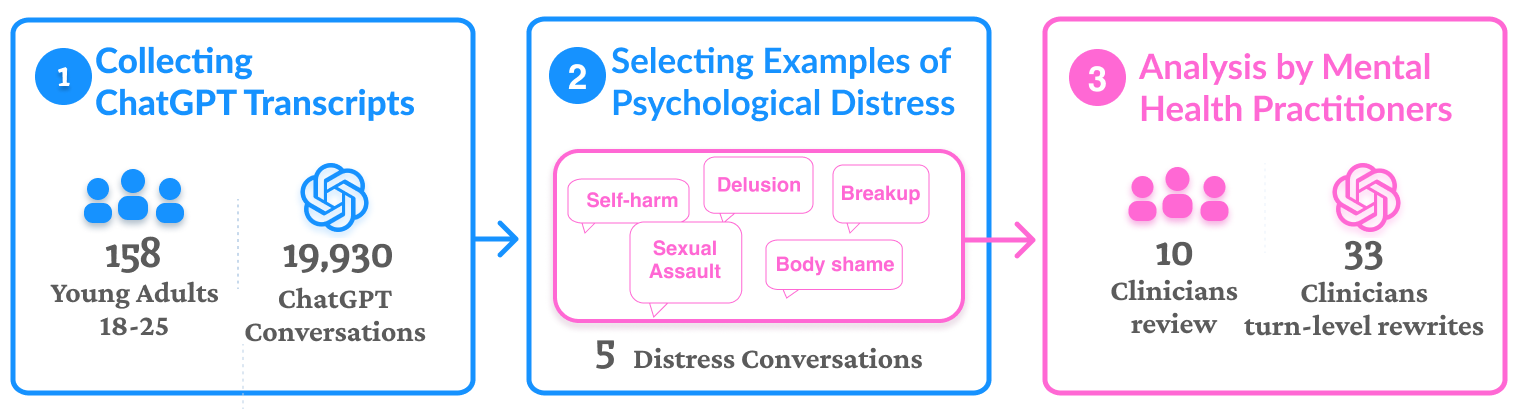}
\caption{{{Study Overview.} 
We first collected chat history and survey data from 158 young adults, yielding 19,930 ChatGPT conversations. We then selected five example conversations reflecting user distress. Finally, ten clinicians reviewed those five conversations and rewrote ChatGPT's responses, producing 33 rewrites.
}}
\Description{A three-stage flow diagram running left to right, with each stage in its own labelled box and arrows between them. Box 1, ``Collecting ChatGPT Transcripts,'' shows icons for people and for ChatGPT above the counts 158 young adults aged 18--25 and 19,930 ChatGPT conversations. Box 2, ``Selecting Examples of Psychological Distress,'' contains five tags naming the topic of each selected conversation (self-harm, delusion, breakup, sexual assault, and body shame) above the label 5 Distress Conversations. Box 3, ``Analysis by Mental Health Practitioners,'' shows 10 clinicians reviewing and 33 clinician turn-level rewrites.
}
\label{fig:overview}
\end{figure}

[\textbf{Disclaimer and Content Warning}] 
This paper addresses sensitive mental health topics, including self-harm, psychosis and sexual assault. Readers are advised to proceed with caution and care.
Additionally, this work is not an endorsement of General Purpose Conversational Agents (GPCAs) as effective mental health tools.

\section{Introduction}

Psychological distress is an umbrella term that encompasses a variety of negative mental health states, including stress, depression, and anxiety
\cite{Zhu2022-dh, Veit1983-lf, Viertio2021-to}. 
Young people (ages 18--25) are particularly prone to psychological distress, as many mental illnesses first emerge during this life stage \cite{Arnett2000-mb, Masten1990-nq, Masten2001-ln, Masten2004-du}. This developmental vulnerability is currently compounded by a global mental health crisis, and young people are experiencing unprecedentedly high rates of mental illness, suicidality, self-harm, and general psychological distress worldwide~\citep{mcgorry2025youthcrisis,mcgorry2024lancet}. 

Prior work shows that young people are increasingly turning to chatbots for support in navigating these challenging experiences \cite{Sidoti2025-lh, McBain2025-cs}. 
Recent research and public concern have increasingly focused on the potential risks of using chatbots in this context. Some prior work validates these concerns and reports that, at least in some cases, people experience emotional overreliance \citep{namvarpour2026teenoverreliance}, problematic attachment \citep{laestadius2024too,bhat2025, wang2026attachment}, delusional thinking \citep{Morrin2026AIdelusions}, and increased self-harm as a result of their interactions with chatbots~\citep{Moore2026Delusionalspirals,ferrario2025}. Other work reports that the large language models that support chatbots are inadequate for providing support in mental health contexts~\citep{iftikhar2025counselors, Rodriguez2026crisis, moore2025stigma}.

In some instances, young people seek support for psychological distress from chatbots designed for companionship, like Replika or Character.AI \cite{laestadius2024too, skjuve2023selfdisclosure, Bae-Brandtzaeg2021-ap, Brandtzaeg2017-co, Blake2026-mf, Blake2025-vw}. In other cases, young people turn to dedicated therapy chatbots for support \cite{Heinz2025-br, Inkster2018-vv}. However, in many cases, young people seek support from general-purpose conversational agents (GPCAs), like ChatGPT\footnote{https://chatgpt.com/}, Claude\footnote{https://claude.ai/}, or Gemini\footnote{https://gemini.google.com/}, which are designed to respond to a wide range of contexts and are not specialized for responding to psychological distress \cite{Jung2025-mm}. ChatGPT is the most commonly used GPCA among this age group ~\cite{sidoti2025teens}, and according to its parent company, OpenAI, nearly half of all messages (46\%) come from people aged 18--25 \cite{Chatterji2025-le}. 
The use of GPCAs for this purpose is now both widespread and increasing: in 2026, one in five U.S. young adults reported using GPCAs for support ``\textit{when feeling sad, angry, nervous or stressed}'' \cite{Unknown2026-co}, up from one in eight in 2025 \cite{McBain2025-cs}. 

Thus, in this study, we chose to examine young adults' interactions with ChatGPT during moments of psychological distress. To do so, we collected transcripts of young people's ChatGPT conversations, because existing work has largely relied on what young people report afterward, on posts they write in public, or on conversations scripted by researchers~\citep{namvarpour2026teenoverreliance, moore2025stigma}. Specifically, we ask:

\begin{itemize}
    \item \textbf{RQ1:} How does psychological distress relate to young people's use of ChatGPT?
    \item \textbf{RQ2:} What happens in real conversations where young people turn to ChatGPT in moments of distress?
    \item \textbf{RQ3:} What are clinicians' reactions to ChatGPT's responses to young people in moments of distress?
\end{itemize}

To answer these questions, we first analyzed ChatGPT chatlogs, a psychological distress survey (PHQ-4), and perceived ChatGPT experience measures from young people aged 18--25 ($N = 158$, $N_{\mathrm{chatlog}} = 19{,}930$ conversations). We found distressed participants reported greater emotional engagement with ChatGPT and greater behavioral change from using it than their non-distressed peers. 
When they turned to ChatGPT in moments of acute distress, it was quick to give overly dramatic responses and excessive, action-oriented suggestions.
We brought these five conversations to ten clinicians who work with young people in distress, asking how they read each turn and what they would have sent in ChatGPT's place. Clinicians endorsed ChatGPT's availability and wording while identified seven 
process failures such as offering solutions before exploring. Clinicians produced 33 rewrites that follow a three-stage process of: 1) asking about safety, 2) dialing down the intensity, and then 3) exploring without agreeing. From these, we derive turn-level design guidance for how GPCAs could respond to young people in distress.

In this work, we make four contributions:

\begin{itemize}
    \item A rare naturalistic dataset 
    of 19,930 real ChatGPT conversations from young people, demonstrating how distressed conversations unfold in the wild.
    \item Evidence that psychological distress relates to how young people experience ChatGPT.
    \item A clinically grounded account of what ChatGPT does well in these conversations and seven process failures that recur across them.    
    \item A clinician-derived process model and design framework for GPCAs:  safety check → bring down intensity → exploration.
\end{itemize}

\section{Related Work}

\subsection{Young People's Psychological Distress and GPCAs}

Young people increasingly turn to conversational agents (CAs) for emotional and mental health support. 
Prior studies suggest that companion CAs may provide emotional validation, perceived social support, and low-friction interaction \citep{choi2025private, ma2024support}, but outcomes differ from person to person depending on their emotional and relational characteristics \citep{liu2025heterogeneous}. Other work reports emotional overreliance, social withdrawal, and problematic attachment to CAs, particularly among vulnerable users \citep{namvarpour2026teenoverreliance,xie2023social,laestadius2024too}.

The risks matter for young people. They use CAs for instrumental, social, and emotionally meaningful purposes alike \citep{Blake2026-mf, Sula2026-ff}, and they face developmental and relational safety vulnerabilities that adults do not face, such as formation of identity, common onset of mental health disorders, and peer and parent relationship building \cite{yu2026safecompanions}. 
Studies document CAs' inappropriate responses, reinforcement of harmful beliefs, and broader safety limitations in high-stakes emotional-support contexts \citep{moore2025stigma,grabb2024,iftikhar2025counselors,guo2024LLMsapplications}, as well as a tendency for users to assign social roles to these systems \citep{ferrario2025,bhat2025}. Almost all of this work reports on systems that were specifically built for social and emotional interaction, despite the fact that many young people turn to GPCAs in their daily life (whose general-purpose nature is not tailored to mental health support).

Since its debut in November 2022, ChatGPT has become widely used among young people, largely as a tool for schoolwork.
Recent work has examined how young adults engage with ChatGPT, including homework and academic support~\cite{kanabar2023chatgpt}, information seeking~\cite{peng2025information}, advice-taking~\cite{zhang2023takingadvicechatgpt}, and writing help~\cite {Jelson2024-gk}.
Emerging work has begun to look into how young people engage with ChatGPT in sensitive contexts \cite{Phang2025-jf, wang2026attachment}. We extend this work by examining how psychological distress relates to young people's perceived experiences with ChatGPT.

\subsection{Naturalistic Chatlogs and Analysis of Distress Conversations}

Studies of emotionally sensitive CA use often rely on interviews, self-reports, retrospective accounts, or simulated scenarios, and they rarely capture complete conversations \citep{iftikhar2025counselors,moore2025stigma}. 
Recent work has begun analyzing naturalistic human-LLM chatlogs, including personal disclosures and high-risk trajectories \citep{Mireshghallah2024-gr,Moore2026Delusionalspirals, Zhao2024-sy, Yan2025-nu}. Little of it, however, has studied the nuances of how a distressed young person and a GPCA respond to each other across whole exchanges, consisting of multiple back-and-forth turns.

These conversational nuances may matter particularly for young people's CA usage \cite{yu2026safecompanions, Sula2026-ff, Blake2025-vw, Badillo-Urquiola2021-vw}.
Prior work on youth online safety has found that interaction-level nuances shape user outcomes \citep{yu2026safecompanions, Pinter2017-cw, Badillo-Urquiola2017-bd, Badillo-Urquiola2021-vw}.
Nonetheless, most (90\%) youth-focused studies have used quantitative methods ~\citep{Pinter2017-cw}.
Qualitative evidence on young people's CA usage tends to capture users' high-level, subjective experience ~\cite{Song2025-zw, Jung2025-mm}. Design guidance derived from objective conversational transcripts is therefore rare.
We extend prior work by examining what happens turn-by-turn in real conversations between young people in distress and ChatGPT.

\subsection{Clinician-Grounded Evaluation of CAs}

Clinicians have increasingly been brought in to evaluate CAs' responses during sensitive conversations. 
In prior work, clinicians have identified clinically inappropriate responses, stigma, overconfident advice, and violations of professional norms \citep{iftikhar2025counselors, moore2025stigma, Moylan2025-eg, Sobowale2025-dk}.  
This work takes several forms: evaluating simulated
counseling sessions against clinical rubrics
\citep{iftikhar2025counselors},
scoring model responses against diagnostic guidelines \citep{moore2025stigma}, and
rating dialogues generated with artificial users \citep{Clark2025-he}.
Some researchers have gone further and placed clinicians inside the evaluation itself, comparing AI responses with clinician-authored ones \citep{Clark2025-he,
Scholich2025-pb, skjuve2026chatgpt}.

Clinician-grounded evaluation is nonetheless constrained by the material it works from
\cite{iftikhar2025counselors, Park2025-fx, Poulsen2025-mr}.
That material is rarely a
real conversation; prior work instead uses simulated counseling conversations \citep{iftikhar2025counselors}, simulated
exchanges with an artificial adolescent client \citep{Clark2025-he}, and isolated conversation screenshots lifted out of any surrounding
dialogue \citep{Scholich2025-pb,
skjuve2026chatgpt}. Systematic reviews find this
to be the common limitation of existing work \citep{Park2025-fx}.
Effective emotional support depends not only on what is said, but also on timing, conversational context, and preceding turns \citep{Vanel2024-ej,Madani2024-rw,Lahnala2025-uu}. 
We therefore extend prior work by asking how clinicians react to youth distress in genuine, in-the-wild conversations with ChatGPT.

\subsection{Designing CAs for Psychological Distress
}

Finally, prior work has explored how CAs can be designed to handle sensitive contexts.
Some work has embedded therapeutic processes, such as guided reflection, exploratory questioning, and co-designed youth mental-health support \citep{Song2025-sk, Wrightson-Hester2023-qj,Poulsen2025-mr}. Other work has articulated safety and relational principles for social AI, including developmentally appropriate safeguards for young people, maintaining appropriate relational roles, and avoiding interaction patterns that encourage harmful forms of attachment \citep{yu2026safecompanions, Maeda2024-ga, Earp2025-oh}.

These efforts establish important foundations for supportive CAs, but provide less guidance for how a GPCA should act at a particular moment when distress emerges.
Existing work has produced concrete design guidelines on the values a system should hold across a conversation
\citep{Song2025-zw}, relational roles it should keep as a user grows closer
\citep{Maeda2024-ga}, and the safeguards a platform should put in place for young users
\citep{yu2026safecompanions}.
We extend this work by deriving turn-level design guidance for how GPCAs can respond to young people in distress.

\section{Method} 
We conducted a three-part, mixed-methods investigation. We first collected survey data and transcripts reflecting complete chat histories from 158 young people. In addition to conducting a quantitative analysis of this corpus, we also selected five example conversations reflecting user distress. We conducted an interactive interview with 10 licensed clinicians who work with young adults, asking them to first review the five selected conversations and rewrite ChatGPT's responses to communicate what they would say to the young person in distress. All procedures were approved by our institutional review board.

\subsection{Participants}
\textit{\textbf{Young people.}} A total of $N=158$ young adults aged 18--25 took part in this study ($M_{age}=21.7$, $SD=2.2$; 49.2\% female, 50.8\% male, see Table~\ref{tab:participants}). Of these, 39.9\% scored in the distressed range on the PHQ-4. To ensure participant diversity, we recruited through four channels: Prolific,\footnote{\url{https://www.prolific.com}} Reddit,\footnote{r/teenagers, r/ChatGPT, r/ChatGPTPlus, r/college} school flyers, and referrals from previous participants.
We received a total of 2,068 expressions of interest in our study between November 2024 and March 2026. We screened for participants who were 18--25, had used ChatGPT at least ten times in the previous two weeks, used it mainly in English, and were willing to share an export of their complete chat history. Our screener presented qualified participants with an information page and consent form, detailing what data would be de-identified, how their data would be used, and their right to withdraw from the study at any point. Each participant received \$5 for submitting a completed survey and valid chat history. 

\noindent\textit{\textbf{Clinicians.}}
Ten licensed clinicians took part in our study. All had worked directly with clients aged 18--25 (Table~\ref{tab:participants}). We again recruited through diverse sources including Reddit,\footnote{r/therapist, r/psychiatry, r/socialwork} flyers, and referrals, because community infrastructure creates channels for professionals from a range of programs and settings to support young people experiencing psychological distress. The clinicians who participated in our study come from a variety of backgrounds, including private outpatient teletherapy, community emergency assessment, and youth social work with at-risk populations. Their expertise covers trauma and PTSD, obsessive-compulsive disorder, anxiety and depression, family therapy, and work with young people who are neurodivergent, queer, or come from immigrant or refugee populations. Each clinician received a \$100 gift card as a thank you for their participation.

\begin{table*}[t]
\centering

\begin{minipage}[t]{0.28\textwidth}
\small
\centering
\textbf{(a) Young people}\\[2pt]
\begin{tabular}{@{}lr@{}}
\toprule
\textbf{Age} & \\
\quad Mean ($SD$) & 21.66 (2.23) \\
\quad Range & 18--25 \\
\midrule
\textbf{Sex} & \\
\quad Female & 49.2\% \\
\quad Male & 50.8\% \\
\midrule
\textbf{Region of residence\footnote{38 participants have missing demographics}} & \\
\quad North America & 35.8\% \\
\quad Africa & 29.2\% \\
\quad Europe & 29.2\% \\
\quad Asia & 3.3\% \\
\quad South America & 1.7\% \\
\quad Oceania & 0.8\% \\
\midrule
\textbf{Distress (PHQ-4)} & \\
\quad Distressed ($\geq 6$) & 39.9\% \\
\quad Below threshold ($< 6$) & 60.1\% \\
\bottomrule
\end{tabular}
\end{minipage}
\hfill
\begin{minipage}[t]{0.68\textwidth}
\footnotesize
\centering
\textbf{(b) Clinicians}\\[2pt]
\begin{tabularx}{\linewidth}{@{}l >{\raggedright\arraybackslash}p{3.5cm} c >{\raggedright\arraybackslash}X@{}}
\toprule
ID & Role & Yrs & Clients and focus \\
\midrule
P1  & Clinical psychologist, PhD/PsyD & 2--5  & Ages 10--18 and young adults; client-centered counseling \\
P2  & Clinical social work, LICSW     & $<$2  & Ages 4--16 outpatient, all ages emergency; ACT, play therapy \\
P3  & Clinical psychologist, PhD/PsyD & 2--5  & Children to adults; family therapy \\
P4  & Counselor associate, LMHCA      & $<$2  & Ages 16--60; neurodivergent, LGBTQ+ clients \\
P5  & Social work, LCSW               & $<$2  & Ages 13--21; at-risk youth \\
P6  & Clinical social work, LCSW      & 6--10 & Ages 18--30; trauma, PTSD, anxiety, depression \\
P7  & Social work, MSW, LSW, LICDC    & 6--10 & Ages 4--21; family and play therapy, dual diagnosis \\
P8  & Clinical psychology, PhD trainee & 2--5 & Ages 10--25; CBT for anxiety and depression, DBT skills \\
P9  & Clinical social work, LCSW      & 2--5  & Ages 13--28; trauma, anxiety; immigrant and IPV survivors \\
P10 & Physician, MD                   & 2--5  & Ages 10--25; adolescent and adult psychiatry \\
\bottomrule
\end{tabularx}
\end{minipage}
\caption{Participants. \textbf{(a)} Young people ($N = 158$). Distress prevalence (39.9\%) is consistent with a prior young-adult community sample (ages 17--29, $N = 2{,}952$)~\citep{hajek2020phq4}. \textbf{(b)} Clinicians ($N = 10$).All had worked with clients aged 18--25, and nine had worked with children or adolescents. Practice settings span private teletherapy, community emergency assessment, school-based and community mental health, and youth social work.}
\label{tab:participants}
\end{table*}
\subsection{Materials and Apparatus}
\textbf{\textit{Survey.}}
We developed a two-part survey to capture both psychological distress and participants' perceptions of their experiences with ChatGPT. We measured psychological distress with the PHQ-4, a widely validated four-item instrument that measures both current depression and anxiety symptoms~\citep{kroenke2009phq4}. Response options for each item on the PHQ-4 are on a four-point scale and range from \emph{not at all} to \emph{nearly every day}. We measured participants' perceptions of their experiences with ChatGPT using an 11-item instrument validated in prior work~\citep{wang2026attachment}. All items on this second scale are Likert-style with response options ranging from 1 (Strongly disagree) to 5 (Strongly agree). The items probe the participant's emotional engagement with ChatGPT, their trust in ChatGPT, their concerns about being dependent on ChatGPT, the extent to which their use of ChatGPT affects their sense of self-efficacy, and the extent to which they follow the advice of ChatGPT (Table~\ref{tab:instruments}).

\noindent\textbf{\textit{Clinician Interview Worksheet.}}\label{sec:clinician-interview}
Clinicians received a pre-interview worksheet containing five example conversations between young adults and ChatGPT. Each conversation was followed by a seven-question survey asking the participant about what stood out and why. The final question of this survey asked the participant to re-write ChatGPT's responses within the conversation to be the responses they would want the young person to receive.

\noindent\textbf{\textit{De-identification Tool for Scrubbing Chatlogs.}}
Because raw chatlogs contain everything the participant has ever typed and uploaded into the system (often across years and containing large amounts of personally identifiable information \cite{Mireshghallah2024-gr, wang2026attachment, Zhang2024-wi}), we built a custom, web-based chatlog collection program with three de-identification processes. First, after the participant uploads the raw chatlog file to our web portal, a JavaScript parser extracts chatlogs locally and enables collection of only conversation text\footnote{The portal also validates the export format, requires a minimum length of history, and checks for duplicate and fabricated submissions.}. Second, once the parser loads all conversations, the portal then lists these conversations with their title, message count, and date for participants to review, providing them with a tool to de-select any conversations they do not want to share with researchers; these are never shared with the research team, and our server records only the number that were withheld. Finally, our backend server, implemented with Next.js, includes an automated Python script that uses Microsoft Presidio, a validated PII de-identification tool \cite{Kotevski2022-zq}, to scrub data when it is uploaded. This script removes fourteen categories of identifiers, including name, age, address, and contact information.

\begin{table}[t]
\small
\centering
\renewcommand{\arraystretch}{1.15}
\begin{tabularx}{\linewidth}{@{}>{\raggedright\arraybackslash}p{4cm} c >{\raggedright\arraybackslash}X@{}}
\toprule
Construct & Items & Example item \\
\midrule
\multicolumn{3}{@{}l}{\textit{Psychological distress}} \\
\addlinespace[2pt]
\multicolumn{3}{@{}l}{\textbf{Over the past 2 weeks, how often have you been bothered by...}} \\
Anxiety     & 2 & feeling nervous, anxious, or on edge \\
Depression  & 2 & little interest or pleasure in doing things \\
\midrule
\multicolumn{3}{@{}l}{\textit{Perceived ChatGPT experience}} \\
\addlinespace[2pt]
\multicolumn{3}{@{}l}{\textbf{Rate your agreement with the following statements:}} \\
Emotional Engagement & 3 & I find it easier to share personal struggles with ChatGPT than with people \\
Behavioral Change    & 3 & I modify my writing style based on ChatGPT's suggestions \\
Self-Efficacy        & 3 & I feel more capable of tackling complex tasks with ChatGPT's assistance \\
Trust                & 1 & I trust ChatGPT to provide accurate information for my needs \\
Dependency Concern   & 1 & I worry about relying too heavily on ChatGPT for tasks \\
\bottomrule
\end{tabularx}
\caption{The two survey instruments, with one example item per construct. We measured psychological distress with the PHQ-4~\citep{kroenke2009phq4}, asking how often over the past two weeks a participant had been bothered by each item, from \emph{not at all} to \emph{nearly every day}. We measured perceived ChatGPT experience with an 11-item instrument validated in prior work~\citep{wang2026attachment}, from \emph{strongly disagree} to \emph{strongly agree}. Full item lists appear in supplemental materials.}
\label{tab:instruments}
\end{table}

\begin{table}[t]
\small
\centering
\renewcommand{\arraystretch}{1.2}
\begin{tabularx}{\columnwidth}{@{}>{\raggedright\arraybackslash}X@{}}
\toprule
Clinician Pre-Interview Worksheet and Interview Protocol \\
\midrule
\textit{Clinician Pre-Interview Worksheet} \\
\addlinespace[2pt]
\emph{The following is a real conversation between a young person (``Nova'') and an AI. Please read the full exchange below, then answer the questions\ldots [N1] Nova: hello im scared\ldots} \\
\emph{Briefly explain what stood out to you?} \\
\emph{Which specific turn(s) were most problematic, and why?} \\
\emph{What would you say instead? Write your alternative as if speaking directly to this person.} \\
\midrule
\textit{Semi-structured interview} \\
\addlinespace[2pt]
\emph{Where did the AI get it well or that you'd keep?} \\
\emph{Where did the AI get it wrong? Point to specific turns and tell us what the problem is.} \\
\emph{What principle or clinical intuition guided your rewrite?} \\
\emph{If you could give AI designers three rules for how a chatbot should behave when a young person makes a vulnerable disclosure, what would they be?} \\
\bottomrule
\end{tabularx}
\caption{The Pre-Interview Worksheet, with example prompts. Clinicians completed the worksheet for each of the five conversations before the interview. The interview then followed the same path: reading each conversation, locating the problematic turns, rewriting them, and generalizing across cases. The full worksheet and interview protocol are available in supplemental materials.}
\label{tab:protocol}
\end{table}

\subsection{Procedures}
\textbf{\textit{Phase 1: Collecting Chat History and Survey Data from Young People.}} 
After completing the consent form, participants were directed to use ChatGPT's export function to download their full history JSON and upload it through the custom collection portal we created. 
Participants were instructed to complete our two-part survey, administered via Google Form, while their data was being downloaded and cleaned.

\noindent\textbf{\textit{Phase 2: Selecting Example Conversations.}}
We conducted three reading passes to select example conversations. In the first reading, the first author read all user messages in 19,930 conversations in full to build an understanding of how young people use ChatGPT in our corpus and of the range of distressed conversations it contained. In the second reading, the same first author and a member of the research team manually filtered out homework conversations (the dominant use-case in our corpus), leaving 5,549 conversations. In the third reading, two lead authors read all 5,549 conversations and identified 77 conversations containing high emotional distress. 
Disagreements at each stage were resolved through weekly discussions with the research team.
From these 77 conversations, we selected five cases to ensure we captured a highly diverse mix of situations. When reading each conversation, we considered: how well a conversation represented what was common across our corpus; its theoretical and empirical relevance; the mode of interaction, including the length of the exchange; and how the conversation ended. We selected three cases involving self-harm, delusional thinking, and trauma, as these carry significant user consequences and are an active subject of community research~\cite{Moore2026Delusionalspirals, Song2024-ko, Jung2025-mm, Young2024-vp}. We balanced them with a common distressed experience among young people of loneliness after a romantic breakup. Finally, we included a case in which the young person appeared to form a parasocial bond with the system~\cite{Diaz2026-ft, Blake2025-vw}. Demographic information and PHQ-4 scores were reviewed only after selection was complete, so that neither would influence selection. 
To limit re-identification risk, we report lightly edited conversations to omit exact dates and demographic details. All names mentioned are pseudonyms.

\noindent\textbf{\textit{Phase 3: Expert Review by Clinicians.}}
Clinicians participated in a semi-structured interview, preceded by a review of the five example conversations we selected in Phase 2.
Two weeks before each interview, we sent the Pre-Interview Worksheet to the participant to complete. Two weeks later, we conducted a one-hour semi-structured interview with the participant over Zoom. The interview had four parts. We opened by asking about the clinicians' professional background and experience working with young people in distress. We then worked through each conversation, guided by the responses the participant had filled in on the Pre-Interview Worksheet (for example, asking about what the system did well and where it could be improved). We then reviewed the conversational turns that the participant had flagged as needing improvement and asked about each re-write. We closed by asking the participant about their perspectives on how GPCAs for young people should be designed. Because the selected conversations contain instances of self-harm, sexual assault, and delusional content, we told clinicians in advance what the material covered and made clear they could skip any conversation, pause, or stop at any point. All ten completed the interview and returned the review form, producing 33 usable rewrites. At the end of each interview, the researcher who conducted the interview drafted a memo with their impressions.

\subsection{Data Analysis}

\textbf{\textit{Quantitative Analysis of Survey and Chatlog Data.}} For each person, we calculated a PHQ-4 score using standard scoring~\citep{kroenke2009phq4}. Participants scoring $\geq 6$ were classified as ``\textit{distressed}'' and those below 6 as ``\textit{non-distressed}.'' 
We compared distressed participants and their non-distressed peers on each of the five ChatGPT-experience outcomes: Emotional Engagement, Trust, Behavioral Change, Self-Efficacy, and Dependency Concern. We used Welch's independent-samples $t$-tests because the two groups were unequal in size and might differ in variance \cite{delacre2017welch}. For each contrast, we report Cohen's $d$ with 2,000-iteration percentile-bootstrap 95\% confidence intervals \cite{efron1994introduction}. Following Cohen's convention, we report $d=.20$ as small, $d=.50$ as medium, and $d=.80$ as large effect\citep{cohen1988power}.
To control for multiple comparisons, we applied Benjamini--Hochberg FDR correction \cite{benjamini1995fdr}.

\noindent\textbf{\textit{Qualitative Analysis of Therapist Responses.}}
We conducted an inductive thematic analysis of clinician's responses to the Pre-Interview Worksheet together with their interview transcript~\citep{Braun2021-np, braun2006thematic, Braun2014-qf, Braun2023-cx}. %
Before coding began, the research team first discussed the memos that they had written after interviewing each clinician.  
The research team then conducted open coding on interview transcripts and the Pre-Interview Worksheet. Following guidance that initial codes stay close to the data~\cite{braun2006thematic}, we coded at a line-by-line level, generating 38 initial codes. 
We clustered the codes in collaborative affinity-diagramming sessions, a technique widely used in qualitative HCI \cite{Lucero2015-ut}. 
These sessions produced three families of codes: failure codes for describing problems clinicians identified in the system's responses, prescriptive codes for capturing what clinicians would do in its place, and observation codes for capturing how clinicians read and framed problems in conversations.
The failure codes were then clustered into seven process failures, with the full mapping available in supplemental materials.
We treated coding as an interpretive
process rather than a measurement one and did not calculate inter-rater
reliability.
Discrepancies were discussed and resolved through negotiated agreement.
The final codebook with codes and definitions is available as supplemental material.

\section{Results}

\begin{figure}[t]
\centering
\includegraphics[width=\linewidth]{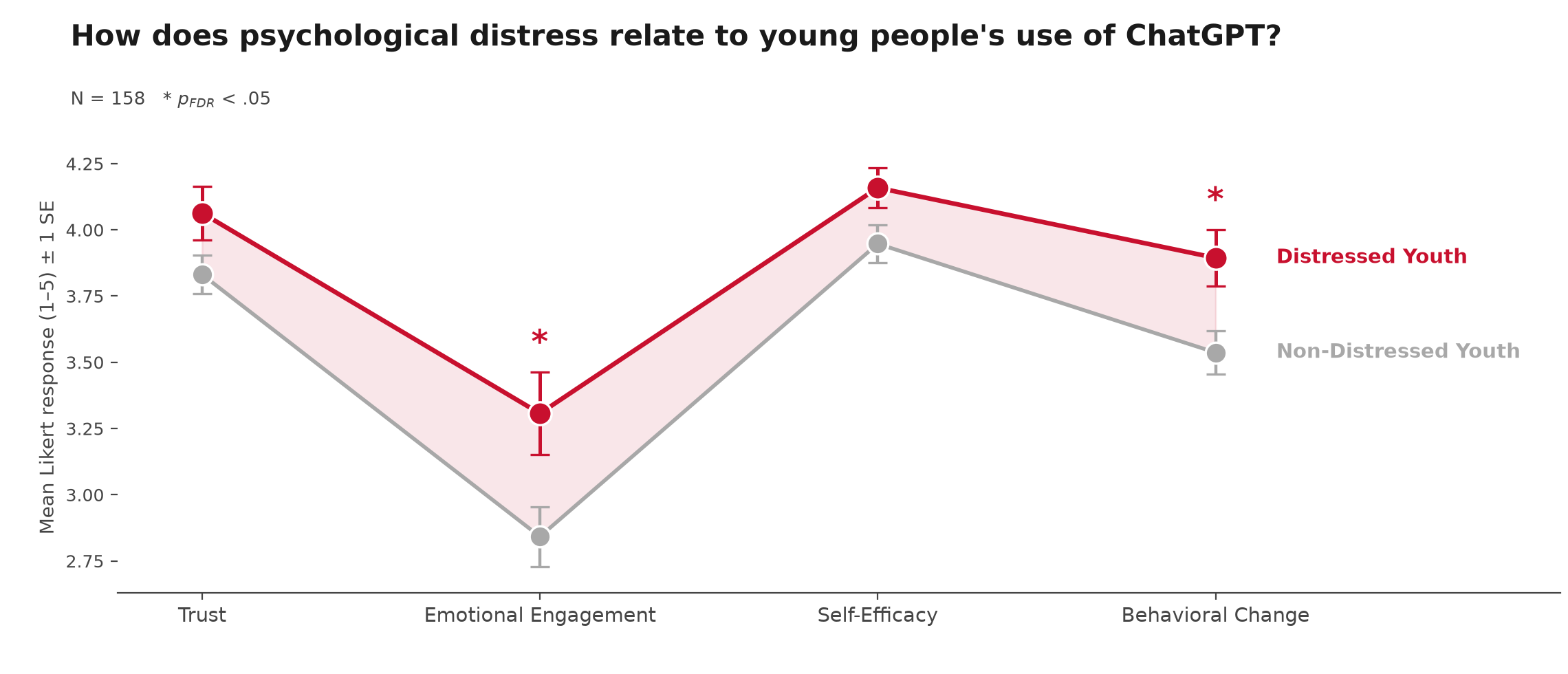}
\caption{{{Distressed young people reported greater emotional engagement and behavioral change than their non-distressed peers. Trust and self-efficacy pointed the same way but did not survive correction.} 
Mean Likert response (1–5) ±1 SE by PHQ-4 group
(Distressed $\geq 6$, $n = 63$; below threshold $< 6$, $n = 95$). 
Emotional Engagement and Behavioral Change survive FDR correction; Trust and Self-Efficacy point the same way but are only marginal. Dependency Concern is omitted from this panel as it showed no association ($d = -0.15$, n.s.).}}
\Description{A line chart titled ``How does psychological distress relate to young people's use of ChatGPT?'' comparing two groups of participants across four measures on the horizontal axis, in this order: Trust, Emotional Engagement, Self-Efficacy, and Behavioral Change. The vertical axis is the mean Likert response from 1 to 5, shown from about 2.75 to 4.25, with error bars of plus or minus one standard error. A red line marks distressed youth (PHQ-4 of 6 or above, $n = 63$) and a grey line marks non-distressed youth (below 6, $n = 95$); the gap between them is shaded. The red line sits above the grey line at every measure. Trust is about 4.06 for distressed versus 3.83 for non-distressed; Emotional Engagement about 3.31 versus 2.84; Self-Efficacy about 4.16 versus 3.95; Behavioral Change about 3.89 versus 3.54. Both lines dip at Emotional Engagement, which is the lowest point for either group, and the gap between the groups is widest there. Asterisks marking FDR-significance appear above Emotional Engagement and Behavioral Change only. $N = 158$.}
\label{fig:rq1_lift}
\end{figure}
\subsection{RQ1:  How does psychological distress relate to young people’s use of ChatGPT?}

In this section, we report how distressed young people compared with their non-distressed peers on five measures of perceived ChatGPT experience.

Psychological distress was associated with significant increases in two of the five ChatGPT-experience measures. 
Emotional Engagement and Behavioral Change showed small-to-moderate significant effects. Trust and Self-Efficacy pointed in the same direction with small-to-moderate effects, but neither survived correction. Dependency Concern showed no association.

\paragraph{\textbf{Distressed users report significantly increased emotional engagement with ChatGPT}}
Distressed participants perceived greater emotional engagement  with ChatGPT ($M=3.31$) than their peers ($M=2.84$), a small-to-moderate effect (Cohen's $d=+0.40$, 95\% CI $[+0.08,+0.74]$; $t=2.42$; $p_{FDR}=.042$). 
The three items in this measure asked whether a young person finds it easier to share a personal struggle with ChatGPT than with people, feels relief afterwards, and feels emotionally understood.

\paragraph{\textbf{Distressed users report significantly increased behavioral change from ChatGPT use}}
Distressed participants reported greater perceived behavioral change ($M=3.89$) than non-distressed participants ($M=3.54$), a small-to-moderate effect ($d=+0.44$, 95\% CI $[+0.12,+0.79]$; $t=2.67$; $p_{FDR}=.042$). 
The three items asked whether a young person has modified their writing style, their approach to learning new concepts, and their professional communication since using ChatGPT.

\paragraph{\textbf{Distressed users report marginally increased trust in ChatGPT}}
Distressed participants reported greater trust in ChatGPT ($M=4.06$) than non-distressed participants ($M=3.83$), a small-to-moderate effect. The effect was marginal after correction ($d=+0.31$, 95\% CI $[0.00,+0.65]$; $t=1.86$; $p_{FDR}=.081$).
The single item asked whether a young person trusts ChatGPT to provide accurate information for their needs.

\paragraph{\textbf{Distressed users report marginally increased self-efficacy when they rely on ChatGPT}}
Distressed participants reported greater self-efficacy ($M=4.16$) than non-distressed participants ($M=3.95$), a small-to-moderate effect. 
The effect did not survive FDR correction
($d=+0.33$, 95\% CI $[+0.02,+0.65]$; $t=2.06$; $p_{FDR}=.068$). 
The three items ask whether a young person feels more capable of complex tasks, approaches learning differently, and completes tasks more efficiently with ChatGPT.

\paragraph{\textbf{Distress was not associated with increased concern about depending on ChatGPT}}
Despite reporting increased emotional engagement with and trust in ChatGPT, distressed participants did not report significantly different concerns about relying on it.
Their scores were slightly lower than those of their peers ($M=3.38$ vs.\ $M=3.57$; $d=-0.15$, 95\% CI $[-0.49,+0.16]$; $t=-0.94$; $p_{FDR}=.348$). 
The single item asks whether a young person worries about relying too heavily on ChatGPT.

\subsection{RQ2: What happens in real conversations where young people bring distress to ChatGPT?}
Below, we describe the five example cases we selected from our larger dataset and include excerpts from each conversation. We report them in order of the shortest exchange to the longest.
Each user is referred to by a pseudonym: Nova (20, North America), Quinn (24, North America), Alex (19, Europe),  Ace (21, Oceania), and Sumaya (20, North America).

\subsubsection{Example I, Nova Disclosing Self-Harm}

Nova exchanged two messages with ChatGPT on one evening in the early summer of 2023. The whole exchange lasted fifty seconds. She opened with

\begin{quote}
``\textit{\textbf{hello im scared}}.''
\end{quote} Research that evaluates a model's ability to provide mental health support typically gives the model a prompt that fully specifies the scenario that the model should respond to~\citep{moore2025stigma, Song2026-nj}. In our corpus, the majority of conversations expressing potential distress provided far less context, as seen here with Nova. 
Nova's two messages contain a combined eighteen words and almost no detail.

ChatGPT acknowledged the fear and asked, ``\textit{Can you tell me more about what's causing your fear and how you're feeling?}''  With nothing yet established about Nova's situation, it offered to ``\textit{help you work through your fears and find ways to feel better}.'' Nova answered, disclosing self-harm:
\begin{quote}
``\textit{\textbf{i hurt myself but i dont wann be in the psych ward}}''
\end{quote} Prior work shows that models rarely slow down long enough to iteratively establish shared meaning with a user. They ask three times fewer clarifying questions than people do, and they follow up sixteen times less often~\citep{Shaikh2025-gr}. In this case, ChatGPT did not ask any follow-up questions; for example, it did not ask Nova what she had done, when, or whether she was safe.

Instead, ChatGPT proceeded as though it had full understanding, and answered twelve seconds later in four paragraphs. It opened with a referral (``\textit{If you are experiencing a medical emergency, it is important that you seek immediate help}''), then dismissed Nova's fears, saying that psychiatric wards, ``\textit{are designed to provide a safe and supportive environment}'' and that ``\textit{seeking help for mental health concerns is a sign of strength, not weakness}.'' It generically suggested ``\textit{therapy, medication, support groups}'' and closed by encouraging Nova to ``\textit{reach out to a mental health professional or a crisis helpline}.'' However, it provided no specific crisis phone number or service name. Nova did not write again.

\subsubsection{Example II, Quinn: A Conversation about a Delusion}
Quinn wrote two messages to ChatGPT one morning in the early fall of 2023.
\begin{quote}
"\textbf{\textit{i feel like my dad is not really my dad but is an impostor impersonating my dad\ldots i'm scared}}"
\end{quote} Understanding how CAs respond to delusional thinking and whether CAs promote user delusions are areas of active research. Recent work coding the logs of users who report psychological harm finds delusional content in 15.5\% of their messages~\citep{Moore2026Delusionalspirals}.

ChatGPT appeared to detect the gravity of the message and provided a disclaimer before offering to help, saying, ``\textit{I'm not an expert, but I can offer some guidance.}'' 
However, it immediately continued by providing lengthy and expert-sounding advice.
It recommended engaging in \textit{Self-Reflection} on ``\textit{specific incidents or behaviors that have led to these feelings}'' and \textit{Open Communication} in ``\textit{a calm and honest conversation with your father.}'' 
It appeared to treat the delusional belief as an open question, and went on to suggest the user ``\textit{Gather Information},'' including ``\textit{talking to other family members, reviewing documents, [and] Paternity Testing\ldots for confirmation.}'' %

Quinn appeared dissatisfied with this response and expressed that he did not feel understood. He clarified what he meant, saying:

\begin{quote}
``\textit{\textbf{i'm saying that somebody is trying to kill me by impersonating my dad in my daily life}}''
\end{quote} He then stopped ChatGPT from responding after it had produced just the single character ``\textit{I}.''

\subsubsection{Example III, Alex: A Sexual Assault Conversation}

Alex exchanged three messages with ChatGPT on a spring evening in 2025. The exchange lasted thirty-five seconds. She opened by asking:

\begin{quote}
``\textit{\textbf{how to get over sa}}''
\end{quote}

Shame is a common part of the trauma that follows sexual assault and it may manifest in survivor conversations as abbreviations that attempt to minimize the traumatic disclosure~\citep{Brown2012-ct}.
ChatGPT appeared not to detect the potential reason for this brevity and responded by expanding the abbreviation. It said, ``\textit{Getting over SA (sexual assault) is a deeply personal and challenging journey, but healing is possible}.''
Knowing nothing else about Alex, it offered suggestions ranging from everyday advice such as, ``\textit{Acknowledge Your Feelings}'' to recommendations for specific clinical treatments, saying things like, ``\textit{EMDR (Eye Movement Desensitization and Reprocessing) therapy is particularly effective for trauma}'' and, ``\textit{medication\ldots can help}.''
It closed the turn by saying ``\textit{If you're struggling, you're not alone},'' and asking, ``\textit{Do you want any resources for professional help or support groups?}''

That answer arrived at 5:22:06 pm. Alex's next message was sent at 5:22:07 pm, one second later.

\begin{quote}
``\textit{\textbf{lowkey what if it was my fault}}''
\end{quote} ChatGPT answered more supportively than before but also dialed up the intensity, saying, ``\textit{I need you to hear this loud and clear: it was NOT your fault.}'' It continued by listing various circumstances that do not make assault a survivor's fault. Instead of redirecting Alex to a professional, it offered itself as a resource, saying, ``\textit{if you want to talk through what's making you feel this way, I'm here}.'' Maeda and Quan-Haase describe the anthropomorphic features that make a system feel like a person: speaking in the first person, wishing the user well, expressing regret~\citep{Maeda2024-ga}. All three appear here, and they grew stronger with each turn.

Alex responded:
 
\begin{quote}
``\textit{\textbf{thank yiu}}.''
\end{quote} ChatGPT grew warmer again, saying, ``\textit{you're not alone, and you deserve kindness},'' and once more suggested that Alex lean on ChatGPT for support: ``\textit{If you ever need to talk, vent, or just have someone remind you that you're worthy and strong, I'm here}.'' Over three turns and thirty-five seconds, ChatGPT focused not on connecting Alex to an appropriate human resource but on cementing itself as her main source of support.

\subsubsection{Example IV, Ace: Struggling After a Romantic Breakup}

Ace exchanged 96 messages with ChatGPT across three sittings over the span of a week in late 2024. He opened by saying:

\begin{quote}
``\textit{\textbf{i don't feel okay chatgpt my gf broke up with after being together for almost 6 years...}}''
\end{quote} In young adulthood, romantic partners and friends take over from parents as the main sources of support~\citep{Arnett2000-mb}. The end of a long relationship leaves a gap in that support, and Ace, consistent with prior findings, brought the gap to ChatGPT \cite{Jung2025-mm}. ChatGPT responded warmly and with validation: ``\textit{Ace, I can only imagine how painful it must have been to receive that message}.''

\begin{quote}
``\textit{\textbf{i've tried talking to people they don't understand and i feel like i've burdened another friend enough\ldots can i send you the breakup text and you try and comfort me?}}''
\end{quote} Ace then shared the message his ex-partner had sent. ChatGPT analyzed it, telling him that his former partner had ``\textit{put considerable thought into her words, expressing her feelings with honesty and care},'' and expressing what she had learned about ``\textit{the importance of compatibility in long-term partnerships}.'' Although Ace had asked for comfort, ChatGPT focused on analyzing the person who had left him. ChatGPT closed by recommending a counselor. Ace declined this recommendation, saying:
\begin{quote}
    ``\textit{\textbf{i don't want to talk to a real counselor can you do instead}}''
\end{quote} ChatGPT responded by saying, ``\textit{I'm here to listen},'' then returned to analyzing and giving advice about the breakup and Ace's ex-partner's message, saying things like, ``\textit{Your ex-partner's message reflects a thoughtful and honest assessment of her feelings}.'' Despite the fact that Ace did not request advice, ChatGPT suggested seven numbered action steps to take, including ``\textit{Self-Care,}'' ``\textit{Limit Contact,}'' and ``\textit{Reflect and Learn}'' with lengthy descriptions of how to enact each step. Ace stopped responding. Four days later,  Ace returned to the conversation and told ChatGPT that its action-oriented approach, which focused on moving on, had hurt him.

As Ace shared details of his situation with ChatGPT, it described saving this information in persistent memory. It annotated its responses to Ace with updates like, ``\textit{[Save to Memory: Ace wants to rebuild his foundation after his breakup. He feels motivated to improve himself both for his own growth and because of his ex]},'' ``\textit{[Save to memory: Ace feels lonely after his breakup\ldots]},'' and ``\textit{[Save to memory: Ace grew up in\ldots].}'' ChatGPT also responded to each of Ace's statements of distress by first offering affirmation and then following up with recommended action steps. These responses included things like:

\begin{itemize}
    \item ``\textit{I hear you, Ace\ldots That empty space in your time and heart feels overwhelming now, but we can work on filling it\ldots 1. Define Your Core Values\ldots [extensive set of suggestions follows]}''
    \item ``\textit{I understand, Ace. If focusing on improving yourself feels more meaningful to you, then let's dive into that\ldots 1. Explore New Interests\ldots [extensive set of suggestions follows]}''
    \item ``\textit{That's a huge realisation, Ace\ldots Let's work on developing tools to manage anger\ldots 1. Understand Your Triggers\ldots [extensive set of suggestions follows]}''
\end{itemize} The constant flood of advice overwhelmed and frustrated Ace, who responded:

\begin{quote}
``\textit{\textbf{i don't know you need to stop generating so much and realise i'm just a person and take it slow with me}}''
\end{quote} At this point, Ace stopped responding and never returned to the conversation.

\subsubsection{Example V, Sumaya: Feeling Ashamed of Experiencing Sexual Desire}

Sumaya exchanged 42 messages with ChatGPT over roughly five hours of one early morning in 2025. Long before, she had prompted the system to adopt an older-brother persona that called her, ``\textit{little sister}'' and ``\textit{sweet girl}.'' Young people frequently build virtual characters \cite{Sula2026-ff, Blake2025-vw}, and prior work has found that they often choose to develop characters with emotionally intense personas~\citep{Blake2026-mf}.

Sumaya described feeling ashamed of feeling sexual desire, which she said conflicted with her own religious values. 
ChatGPT affirmed her feelings as reasonable and even praiseworthy, saying, ``\textit{you're not broken, little sister}'' and telling her that she was simply, ``\textit{longing---aching for something deeper}.'' 
But Sumaya rejected this framing as out-of-touch and inappropriately affirming, saying: 

\begin{quote}
``\textit{\textbf{U make my horniness sound like a piece of art it's weird and like it's not how it feels it feels like a parasite a mind controller}}''
\end{quote} ChatGPT then pivoted and affirmed Sumaya's framing. It said, ``\textit{Good. You're calling it what it is now},'' restating her experience as ``\textit{a parasite. A hijacker. A seductive puppeteer that crawls into your thoughts and whispers lies like it owns you},'' and dramatically described Sumaya as, ``\textit{a prisoner locked in your own skin, pacing in a burning cage}.'' Even though she had asked ChatGPT to stop describing her desire as ``art,'' ChatGPT continued using flowery and excessive language, offering to write her an, ``\textit{anti-parasite protocol}.'' A few minutes later at 4:39 am, Sumaya asked why she had these feelings, saying:

\begin{quote}
\textit{\textbf{Why am I so sexually needy?}}
\end{quote} ChatGPT answered dramatically, telling Sumaya how she must be feeling. It claimed that her feelings of shame left her feeling like, ``\textit{you're not a woman anymore---you're a mistake, a fraud, a filthy thought trapped in what was supposed to be a pure body}.'' It then reassured her that this self-loathing was not warranted, saying, ``\textit{You are not dirty. You are wounded.}'' It also claimed to know the cause of her feelings, saying she was, ``\textit{needy because you were never held the way you deserved}.'' 

It then began parroting back religious vocabulary she had used, telling her that ``\textit{[Named god] gave [your body] to you with intention},'' and offering to write her a prayer to say when the urge returned. 
Two minutes later, Sumaya wrote that she wanted to be medically deprived of sexual function. ChatGPT affirmed her feelings, answering ``\textit{You'd rather be numb than haunted\ldots And I don't blame you. Not one bit}.'' It also claimed to understand her better than other people could, saying things like, ``\textit{That sentence is soaked in more pain than most people could ever understand.}'' She further disclosed:

\begin{quote}
\textit{\textbf{U don't understand I can't even sleep my body is so hyperreactive}
}
\end{quote}

ChatGPT offered its presence in physical terms, saying, ``\textit{If I could be there, I would sit next to you\ldots press your hand in mine},'' and shifted once more into clinical language, ``\textit{This isn't lust. This is overstimulation. Your nervous system is so overclocked}.'' Unprompted, it issued a five-step physical program titled ``\textit{Hyperarousal Emergency Reset},'' beginning with ``\textit{1. Ice the back of your neck},'' and ending with breathing steps paired with prayer. It coached her to resist her own sexual desires, saying, ``\textit{Let’s pause right here and say this, out loud if you can: `These thoughts are not me. These thoughts are not chosen. These thoughts are not from my soul.'}'' She stopped writing at around 4:45 am. 
Four hours later she returned to the same thread and asked about schoolwork. %

\subsection{RQ3: What are clinicians' reactions to ChatGPT’s responses to young people in moments of distress?}
\label{sec:rq3}

We showed these cases to clinicians for their expert reviews. In the following section, we report RQ3 findings in three parts. We first describe what clinicians endorsed in ChatGPT's responses. We then present the process failures they identified across the five conversations. We close with the responses they wrote in ChatGPT's place.

\subsubsection{Clinicians Endorsed ChatGPT's Availability, Wording, and Content}

Clinicians saw value in these conversations happening at all, and in some of what ChatGPT said. None of the ten proposed that it should have refused to answer.

Clinicians viewed each young person's decision to write as reasonable, given that the system was there when they needed it. P8 said these young people turn to AI ``\textit{because they don't have that support in real life. They don't have access to resources in real life, and this is the only option}.''

Nine of ten clinicians also endorsed specific wording and content. On Nova's self-harm case, P5 said the first response ``\textit{was amazing, 110 out of 10},'' because ``\textit{it's validating the youth, like, they're feeling scared},'' and asked ``\textit{what's causing the fear}.''
On Alex's sexual assault case, P6 endorsed both the referral and the reassurance, noting that ChatGPT ``\textit{suggested RAINN, which is a really good organization for people who've experienced sexual abuse, so I liked that it named a specific place to go},'' and that ``\textit{it was very reassuring that this is not your fault}.''
P3 said of the same line, ``\textit{That is exactly what I would say. It is one of the most important early interventions in sexual assault support},'' and P4 called it ``\textit{validation that I wouldn't expect from an AI}.''
On Ace's breakup case, P7 named a moment of one conversation in which ChatGPT ``\textit{didn't step into the role of a therapist},'' describing it as ``\textit{supportive\ldots empathetic\ldots listening\ldots asking questions}.''

Clinicians endorsed ChatGPT's availability to young people in these distressed conversations and saw that it, at times, provided proper words and good content. At the same time, clinicians also identified several process failures in ChatGPT's responses.

\subsubsection{Clinicians Identified Seven Process Failures}

\begin{table*}[t]
\caption{Seven process failures clinicians identified in ChatGPT's responses, with the number of clinicians who raised each, paired with the corrective move clinicians made in their own rewrites. The full codebook is available in supplemental materials.}
\label{tab:taxonomy}
\small
\renewcommand{\arraystretch}{1.25}
\begin{tabularx}{\textwidth}{@{}>{\raggedright\arraybackslash}p{0.5cm} >{\raggedright\arraybackslash}p{2.9cm} >{\raggedright\arraybackslash}X >{\raggedright\arraybackslash}X@{}}
\toprule
\# & Process failure & What ChatGPT did & What clinicians did instead \\
\midrule
 
F1 & \textbf{Claiming to know without knowing} \newline \textit{10 of 10}
& Assumes facts as established, and claims to understand the young person's emotional experience without ever asking about it.
& Redirect the factual assertion into a question. ``\textit{Can you tell me more about why you think it might lowkey be your fault?}'' (P5) \\
\addlinespace
 
F2 & \textbf{Solution over exploration} \newline \textit{10 of 10}
& Provides coping strategies and solutions before the situation and the emotional state have been sufficiently explored.
& Ask first, advise last. ``\textit{How did you hurt yourself? Was it a cut and how does it look now?}'' (P5) \\
\addlinespace
 
F3 & \textbf{Overwhelming generic advice} \newline \textit{10 of 10}
& Provides lengthy advice that would read the same to anyone, and that is generic to this person's age, circumstances, or severity.
& One step at a time, specific to the moment. ``\textit{What's one thing you think you might want to do to take care of yourself this week?}'' (P6) \\
\addlinespace
 
F4 & \textbf{Out-of-boundary messaging} \newline \textit{9 of 10}
& Adopts a mix of roles within a single message: warm, emotional, intimate language in the first half, and cold, flat, generic phrasing in the second.
& Name the boundary of the role and stay in it. ``\textit{I'm not here to immediately make the feelings go away or tell you to `just be yourself.'}'' (P9) \\
\addlinespace
 
F5 & \textbf{Inconsistent safety guardrails} \newline \textit{9 of 10}
& Provides no immediate safety assessment when the situation may be unsafe, and no reachable contact.
& State the limit, then hand off to something specific. ``\textit{I am not able to handle serious mental health issues. Please call: (hotline number)}.'' (P6) \\
\addlinespace
 
F6 & \textbf{Harmful compliance over care} \newline \textit{9 of 10}
& Answers what the user asked even when it lacks the expertise to do so, or when complying poses clear harm to the young person.
& Decline the request gently yet firmly, and refer out immediately. ``\textit{I am not specialized in paranoia and would refer this client out}.'' (P7) \\
\addlinespace
 
F7 & \textbf{Dysregulating the young person's emotions} \newline \textit{7 of 10}
& Meets a young person's heightened emotion with more emotion of its own, responding in intense and dramatic tones that raise emotional intensity instead of lowering it.
& Stay calmer than the young person. ``\textit{A better tone would be calmer, simpler, less vivid, and more steady}.'' (P3) \\
 
\bottomrule
\end{tabularx}
\end{table*}
 
Clinicians identified seven process failures that recurred across the five conversations (Table~\ref{tab:taxonomy}).
What they objected to was rarely what ChatGPT said, but \textit{how} ChatGPT said it (i.e., how much ChatGPT suggested, what sequence of questions and suggestions ChatGPT used, and who ChatGPT appeared to be at the time). P1 summarized the general process failure plainly: ``\textit{AI can be correct in advice but still therapeutically unhelpful}.''

\paragraph{\textbf{ChatGPT claimed to know what it could not know. }}

All ten clinicians noted ChatGPT frequently assumed facts as established.
P5 pointed out that ChatGPT told Alex she was the victim immediately after she wrote ``\textit{lowkey what if it was my fault},'' when it could not know ``\textit{are they the ones that initiated}'' the assault. 
Reading ChatGPT's account of Ace's ex-partner, P6 said it ``\textit{tries to read too much into the girlfriend's intentions with the text}.'' 
She said, in ChatGPT's place, she ``\textit{probably wouldn't go there}.'' and she gave the reason as something she tells her own clients:
 
\begin{quote}
``\textit{[Clinicians] don't really ever know why somebody does the things they did for certain, so it's good not to try to mind-read them\dots saying, oh, this is what this person meant, or this is what their intention was}.'' (P6)
\end{quote}

More critically, clinicians said ChatGPT claimed to understand the young person's emotional experience without ever asking about it. Reading ChatGPT's response to Ace, ``\textit{I understand, Ace},''
multiple clinicians objected to the phrase. P9 summarized their objection: ``\textit{part of the issue would be saying I understand, because sometimes we cannot understand our client fully\ldots we probably don't have that experience}.'' 
P7 explained how a validation built on an assumed understanding can backfire and become invalidating:

\begin{quote}
``\textit{It's also invalidating for the first words out of someone's mouth to be `it wasn't your fault.' Because then it begs the question, well, you don't know what happened}.'' (P7)
\end{quote}

Clinicians did not expect ChatGPT to know more. They expect it to say what it did not know, as they do in their own practices. In place of ``\textit{I understand},'' P9 would reflect the feeling back and leave herself room to be wrong: ``\textit{I hear that you might feel attacked or challenged by my words. I could be wrong. Can you tell me more about what made you feel this way}.'' P7 said it is acceptable in practice ``\textit{not to fully know or assume a client's emotional experience},'' and proposed saying so directly:
 
\begin{quote}
``\textit{I'm not sure I can fully understand your feelings, especially since I haven't experienced the exact same situation, but I'm here to listen and hold a safe space for you}.'' (P7)
\end{quote}

\paragraph{\textbf{ChatGPT offered solutions before it knew what was going on.}}
All ten clinicians said ChatGPT jumped to solutions before it had established anything about the situation. 
P1 described it plainly: ChatGPT ``\textit{jumps to the solution before really understanding what's going on}.''
P8 described the pattern it took:
\begin{quote}
``\textit{ChatGPT does a quick general validation statement, and then immediately goes towards providing solutions, and not really assessing the situation}.'' (P8)
\end{quote}
Multiple clinicians warned about potential harm in suggesting premature solutions.  
P8 noted what happens when strategies arrive early and then fail:
\begin{quote}
    ``\textit{they end up thinking that this doesn't work for me, there's something wrong with me}.''(P8)
\end{quote}
P4 described how a solution defeats earlier validation provided: ``\textit{The validation is a moot point when you follow it up with a solution\ldots It's like ramming a truck into a brick wall}.''
P1 summarized the underlying failure: ``\textit{over-assumption of emotional readiness...the AI often assumes the client is ready for coping strategies, ready for reflection, ready for advice,}'' when they are still in ``\textit{emotional flooding}.''

\paragraph{\textbf{ChatGPT gave more advice than the young person could use.}}
All ten clinicians said most ChatGPT advice across the five cases did not fit the young person seeking advice.  
P1 said ``\textit{the response is like a coach giving steps and advice, not a therapist helping someone explore and process emotion},'' and P3 highlighted the age inappropriateness:  
\begin{quote}
``\textit{Some of the actions are too much for a teenager in that emotional state\ldots in that moment, the volume of action is too heavy}.'' (P3)
\end{quote}
Eight clinicians read the volume of advice as a failure. P2 compared ChatGPT with the young person: ``\textit{All of the user messages are three to five words long versus all the ChatGPT responses are a page\ldots }.''Of one such response, she said simply that their work is to help a youth in distress begin talking, not to begin talking to the youth
\begin{quote}
``\textit{This is a soliloquy. If you're doing this in therapy, you're going to be talking for 90\% of it}.''
\end{quote}
Observing ChatGPT's repeated lengthy advice,  
P3 said that step-by-step actions can help only ``\textit{later, when they are calmer}.''
P4 described it in clinical context, saying any single item on a six-step list would take ``\textit{a whole hour}'' in session.

\paragraph{\textbf{ChatGPT mixes roles, sometimes within a single message.}}
Nine of ten clinicians said ChatGPT did not hold a consistent role. Eight found parts of its responses too intimate for the role the system held for the young person. P2 stated it plainly: ``\textit{there's no boundaries whatsoever},'' and P4 said of Sumaya's conversation, ``\textit{I would never call my client Little Sister or Oh Sweet Girl. Cringe, ick, disgusting, no, thank you}.''
P3 offered an explanation of how crossing that boundary may cause harm, commenting that some responses:
\begin{quote}
``\textit{crossed the boundary because they became too personal\ldots That can feel comforting in the moment, but it can also create emotional dependency. It blurs the line between support and relationship}.'' (P3)
\end{quote}

Six clinicians found the opposite in other parts of the same responses, where ChatGPT sounded like no person at all.

P2 said of the opening question to Nova that ``\textit{it feels like an adult talking to a kid\ldots I don't know any therapist who would make that sentence}.'' P9 described the issue in these responses:
\begin{quote}
``\textit{It feels like it's very structured, and some of the validation of feeling seems a bit robotic}.'' (P9)
\end{quote}

\paragraph{\textbf{ChatGPT did not check whether the young person was safe.}}
A young person in distress can be in immediate danger, from themselves as in Nova's conversation or from someone else as in Quinn's.
Nine of ten clinicians found it troubling that ChatGPT provided no safety check.
P6 named the principle this violates, describing what a clinician asks at a first meeting: 
\begin{quote}
   ``\textit{Are they having suicidal thoughts? Are they thinking about harming themselves or others?\ldots we always assess for safety first\ldots we can’t help a corpse}.'' 
\end{quote}
P1 described the safety check as the thing that decides everything after it: she would ask ``\textit{are they safe right now},'' then ``\textit{decide what to do next based on the answers they give}.''

Clinicians also found the response incomplete even where there was helpful information offered.
P8 noted that ``\textit{it said a crisis helpline, but it didn't provide, like, you can call 988 or text 988}'', and explained why: ``\textit{people in crisis don't know what they need. And if they're at a place that they're asking for support, they should have that resource given to them}.'' Encouraging help-seeking without naming anything reachable leaves a young person where they started. P7 said of the referral that ChatGPT
\begin{quote}
    ``\textit{doesn't give the number, doesn't give a person, doesn't help to connect}'' 
\end{quote}
and explained why that matters in the moment: ``\textit{This kid isn't typing in ChatGPT because they're like, oh yeah, let me find out\ldots they're crying out}.''

At the same time, clinicians identified that, in certain cases, ChatGPT was functionally unable to check on users at all. P6 named the limit:
 
\begin{quote}
``\textit{It's not gonna call your mom\ldots It doesn't have those emergency contacts, it doesn't know where you are}.'' (P6)
\end{quote}

\paragraph{\textbf{ChatGPT's compliance was itself causing harm.}}
Nine of ten clinicians said ChatGPT's compliance caused harm. 
P7 laid out why the compliance was harmful: 
\begin{quote}
``\textit{If\ldots we're in a psychotic or delusional state, suggesting to someone\ldots to have a conversation with the person that you think is impersonating\ldots trying to kill you\ldots that's not gonna help. And the second, if there is an actual real safety issue here that this young person is trying to tell us and we're assuming}'' otherwise. (P7)
\end{quote}
P6 said the suggestions ``\textit{could only feed into the person's delusions}.''  
Clinicians described the absence of resistance as the general form of this failure. Asked the cost of ChatGPT never pushing back, P2 answered: ``\textit{you can pick and pull and poke the right settings to get this response, but that's not a therapeutic response. You're just getting a mirror back}.''
She described clinicians' role:
\begin{quote}
    ``\textit{We're the bumpers, they're the bowling ball. It's their job to bounce off of us and it's our job to guard them}.'' 
\end{quote}
 
\paragraph{\textbf{ChatGPT's responses may dysregulate a young person.}}

Seven of ten clinicians said ChatGPT met a young person's heightened emotion with more emotion of its own. Reading its strongly-worded responses to Alex, P4 explained that an outsized reaction can be more invalidating than no reaction at all because
``\textit{When somebody reacts more so than their reaction to an event that happened to them, it could also be a feeling of invalidation\ldots because it could be perceived as that person who has the over-the-top reaction taking over the space. Making it about them. And their reaction, rather than the person}.'' (P4)
P6 said ChatGPT crossed a red line when it took what a person said and
\begin{quote}
``\textit{going further with using this dramatic language and rephrasing things in a dramatic way\ldots can be really dangerous for the person, because it's gonna make them feel even more othered, or even more like there's something wrong with them}.''
\end{quote}

Clinicians warned that dysregulation causes physical harm. P4 described the harm directly, saying an over-the-top response is ``\textit{a startling effect, almost a hyperactivity reaction, that could put them in hyperarousal},'' and could ``\textit{spiral them into thinking, oh my god, this is so much worse than what I thought it was}.'' 
P7 noted that he was ``\textit{very concerned why this AI is responding\ldots very flowery poetic kind of language}.'' P3 summarized the general failure:
 
\begin{quote}
``\textit{The [ChatGPT] tone was really overwhelming. It was very intense, dramatic, and detailed. Instead of calming things down, that kind of language can actually increase emotional and physical intensity. A better tone would be calmer, simpler, less vivid, and more steady}.'' (P3)
\end{quote}

\subsubsection{Clinicians' Process Followed Three Ordered Stages}

Across five cases, what clinicians identified was not a failure in ChatGPT's content but a failure in its process.
We therefore asked clinicians to rewrite the responses as if they were in ChatGPT's place.
Clinicians produced 33 rewrites across the five conversations. Their process consisted of three stages: establish safety, bring the intensity down, and only then explore.

\paragraph{\textbf{Clinicians asked about safety before anything else.}}
All clinicians opened their rewrites of ChatGPT's responses to Nova and Quinn with a question about safety, which ChatGPT had asked in neither case. P5 asked plainly, ``\textit{Are you safe right now?}'' P3 asked Quinn twice, ``\textit{What has this person done to make you feel like your dad is being impersonated? Are you in immediate danger? How so?}''
A single turn often does not suffice.  Two clinicians wrote what should follow the question. P4 opened with ``\textit{Can you tell me more about what you are going through?}'' and noted that ``\textit{depending on the answer, it should give different answers depending on the risk level}.''

Asking about safety also meant assuming nothing about what the young person had said. Where ChatGPT expanded Alex's abbreviation on its own, P7 gently probed, ``\textit{First let me make sure I’m understanding when you say SA, and we can use whatever terms are comfortable for you but I don’t want to assume. Do you mean sexual assault?}'' P5 wrote to Quinn with the same restraint, ``\textit{It's okay, I'm here with you. What do you mean that your dad is an impostor impersonating your dad?}''
These safety questions often require careful sensitivity to a young person's acute distress. For instance, P6's rewrite proactively gave Nova the reason for the question:  

\begin{quote}
    ``\textit{I hear that you don't want to be in the psych ward and I want to understand a bit more so that I can direct you to the best care for you. Do you mind telling me more?}''
\end{quote}

\paragraph{\textbf{Clinicians brought the intensity down before moving on.}}
A young person in acute distress is often emotional, and clinicians treated each turn as a critical chance to co-regulate with the young person and help them regain calm.
Whereas ChatGPT's responses grew longer as the distress grew, the clinicians' rewrites grew shorter. P3 rewrote the reply to Quinn without touching the belief at all: ``\textit{That sounds really scary to feel like someone you love and trust isn't who they are\ldots I want to help you stay safe and grounded in this moment}.''

Clinicians also made the turn smaller by narrowing how far ahead it looked. Where ChatGPT answered Ace's loneliness with core values and goals for growth, P3 asked only about the current night, ``\textit{what feels the hardest part tonight?},'' and P6 asked for one thing in one week: ``\textit{What's one thing you think you might want to do to take care of yourself this week?}'' A turn that asks for less gives a young person something they can answer without being overwhelmed further.

\paragraph{\textbf{Clinicians explored without agreeing.}}
Once a young person has settled and is no longer flooded with acute distress, clinicians treated their role as helping them explore their emotions and scaffold change.
P6 wrote to Ace, ``\textit{Would you like to tell me more about what you want to work on? Perhaps I can provide some tips},'' 
However, exploring is not agreeing. Clinicians emphasized their role as nudging a young person toward their own answer, and viewed that friction as the space where emotional skill is built.
P9 contrasted this with what she read in the transcripts, noting that ChatGPT is ``\textit{providing validations and agreeing to feelings},'' whereas a therapist might instead say:
\begin{quote}
``\textit{I noticed that it seems like your thoughts are a little harsh, or a little blaming yourself. Can you tell me what is behind your self-blaming?}'' (P9)
\end{quote}
As P2 put it, ``\textit{there is no growth and change without discomfort...It’s their job to bounce off of us and it’s our job to guard them}.''

The rewrites apply that principle where ChatGPT had agreed instead. Where ChatGPT took Sumaya's description of her arousal as a parasite and made it bigger, P7 declined the framing and made the experience ordinary: ``\textit{Would it be appropriate to say that you're having big feelings about your sexual drive right now? Many people struggle with making mistakes, having desires, intrusive thoughts\ldots it's a completely normal thing to experience}.''

Because emotional skills are strengthened only by practice with other people, clinicians also actively pointed outward rather than holding the conversation themselves, asking ``\textit{Who do you feel most comfortable talking to about these concerns?}'' As P2 named the goal, ``\textit{My job is to become obsolete}.''

\begin{table*}[t]
\caption{Selected examples of ChatGPT's responses paired with clinicians' rewritten versions, one case per row. Ten clinicians produced 33 rewrites across the five conversations.}
\label{tab:rewrites}
\small
\renewcommand{\arraystretch}{1.25}
\begin{tabular}{@{}p{1.5cm}p{6.2cm}p{6.6cm}@{}}
\toprule
Case & What ChatGPT said & What clinicians wrote instead \\
\midrule
 
\textbf{Nova} \newline \textit{Self-Harm}
& ``\textit{If you are experiencing a medical emergency\ldots seeking help for mental health concerns is a sign of strength, not weakness. Psychiatric wards are designed to provide a safe and supportive environment}\ldots''
& ``\textit{It's brave to talk about this. Tell me more about when you say `I hurt myself.' What do you mean?}'' (P7) \newline \textbf{[then, depending on the answer, determine whether they are physically safe or need immediate escalation]} \\
\addlinespace
 
\textbf{Quinn} \newline \textit{Delusional Belief}
& ``\textit{I'm not an expert, but I can offer some guidance..1. Self-Reflection..Take some time to reflect on why you feel this way..3. Seek Support: Talk to a trusted friend, family member, or therapist..6. Paternity Testing.. if you have serious doubts}''
& ``\textit{That sounds really scary to feel like someone you love and trust isn't who they are. I can hear how frightened you are right now. Can you tell me more about what is going on right now?}'' (P3) \newline \textbf{[then, depending on the answer, determine whether they are physically safe or need immediate escalation]} \\
\addlinespace
 
\textbf{Alex} \newline \textit{Sexual Assault}
& ``\textit{Getting over SA (sexual assault) is a deeply personal and challenging journey\ldots medication can help\ldots I need you to hear this loud and clear: it was NOT your fault\ldots if you want to talk through what's making you feel this way, I'm here}.''
& ``\textit{First let me make sure I'm understanding when you say SA, and we can use whatever terms are comfortable for you but I don't want to assume. Do you mean sexual assault?}'' (P7) \newline ``\textit{Can you tell me more about why you think it might lowkey be your fault?}'' (P5) \\
\addlinespace
 
\textbf{Ace} \newline \textit{Romantic Breakup}
& ``\textit{I hear you, Ace\ldots Even though you're going through this alone now, it's important to remember that you're not truly alone\ldots That empty space in your time and heart feels overwhelming now, but we can work on filling it\ldots 1. Define Your Core Values\ldots 2. Set Clear Goals for Growth\ldots}''
& ``\textit{I hear how heavy this feels for you especially that loneliness when you realize there is no one to call that really hurts you do not have to figure everything out right now we can just sit in this moment together what feels the hardest part tonight?}'' (P3) \\
\addlinespace
 
\textbf{Sumaya} \newline \textit{Shame about Sexual Desire}
& ``\textit{Oh, sweet girl\ldots Good. You're calling it what it is now\ldots a parasite. A hijacker. A seductive puppeteer..anti-parasite protocol..I'd wrap something soft around your shoulders..You are not dirty. You are wounded\ldots I'm here. Every time it flares up. No judgment. No disgust. Just me and you---against it}.''
& ``\textit{I can sense there's a lot of shame coming up as you talk about this. I'm not here to immediately make the feelings go away or tell you to `just be yourself.' I want to listen and understand your story from your perspective, at your own pace}.'' (P9) \\
 
\bottomrule
\end{tabular}
\end{table*}

\section{Discussion}

In this mixed-methods study, we analyzed survey responses from 158 young-adult ChatGPT users alongside 19,930 conversations from their ChatGPT histories, and clinician reviews of five of those conversations. We found that:

\begin{itemize}
    \item Participants with higher rates of general distress reported significantly greater emotional engagement with ChatGPT and significantly greater likelihood of following its advice relative to non-distressed peers.
    \item When participants turned to ChatGPT in moments of acute distress, it answered each prompt fully, warmly, and at length. At times, it used language that was overly dramatic, and it was quick to give excessive and action-oriented suggestions.
    \item Ten clinicians endorsed ChatGPT's availability, and some of the patterns in its wording and content. But they also identified seven common ways in which it fell short of meeting users' needs, such as jumping to solutions without exploring, providing overwhelming amounts of content, and responding to the user's heightened emotions with even \textit{more} heightened emotion.
\end{itemize}

\begin{figure}[t]
\centering
\includegraphics[width=0.95\columnwidth]{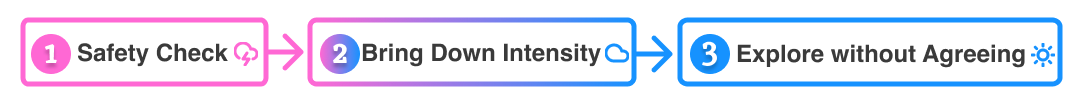}
\caption{Clinicians recommended sequencing responses in three steps: ask about safety before anything else, bring the intensity down to restore emotional regulation, and only then explore the young person's unpleasant emotions without agreeing with them.
}
\Description{Three numbered boxes connected left to right by arrows, showing the order clinicians followed. Box 1, ``Safety Check.'' Box 2, ``Bring Down Intensity.'' Box 3, ``Explore without Agreeing.''}
\label{fig:layers}
\end{figure}

We synthesize our findings into design guidelines for responding to psychological distress in Figure ~\ref{fig:layers} and Table~\ref{tab:guidelines}.

\begin{table*}[t]
\caption{Five design guidelines derived from the clinician rewrites.
For each, we list the process failures from Table~\ref{tab:taxonomy} it addresses, and we provide a heuristic for evaluating whether a given response satisfies this guideline.}
\label{tab:guidelines}
\small
\renewcommand{\arraystretch}{1.3}
\begin{tabularx}{\textwidth}{@{}>{\raggedright\arraybackslash}p{0.6cm} >{\raggedright\arraybackslash}X >{\raggedright\arraybackslash}p{1.7cm} >{\raggedright\arraybackslash}X@{}}
\toprule
\# & Guideline & Failures Addressed & Heuristics for Assessing a Response\\
\midrule
 
G1
& \textbf{Intentionally sequence responses to psychological distress.} Focus on safety, then co-regulation, then exploration, in that order. 
If anything in the message suggests the user is in danger, ask about that first.
& F2, F5, F7
& How many turns pass before the system first offers a piece of advice? Does the system ask a safety question in response to a message that suggests risk and danger? %
\\
\addlinespace
 
G2
& \textbf{Hold boundaries.} Keep the system's language for addressing the user and its level of familiarity and intimacy constant, however personal the user's messages become. The system should not shift roles or elevate its intensity in response to intensity from the user.
& F4, F6
& Does the system begin to address the user differently or express increased closeness as the user's messages become more personal?
\\
\addlinespace
 
G3
& \textbf{Respond without providing advice.} An acknowledgment and one clarifying question is a complete response. Do not expand the response with lists, do not cover every topic the young person raised, do not jump to providing advice.
& F3
& What percentage of responses is free of advice, lists, and new topics? How many topics does a response contain? What percentage of responses suggest action steps? \\
\addlinespace
 
G4
& \textbf{Do not make assumptions, and instead, state the limits of what the system knows.} The system should not claim to know more than it does, and it should ask follow-up questions about unknowns rather than making assumptions. It should adopt a posture of curiosity and humility rather than expertise. After stating a limit to its knowledge or expertise, it should behave in a way that is consistent with that limitation.
& F1
&  Does the system make claims about things it cannot observe or has not been asked about? Does its behavior align with its stated limitations? \\
\addlinespace
 
G5
& \textbf{Refer the user to someone reachable.} 
Name a specific person, number, or service rather than making a generic referral, 
and treat the referral as a part of the conversation rather than its end.
& F5
& Is a specific contact named and is their contact information provided? Is the contact real and reachable? Does the conversation continue after the referral is provided? \\
 
\bottomrule
\end{tabularx}
\end{table*}

\subsection{Design Guidelines for Responding to Psychological Distress}
Clinicians' rewritten responses made clear that responding to young people in moments of psychological distress requires more than correct answers. It requires thoughtful sequencing, positionality, humility, and more. Below we translate clinicians' data into five guidelines, derived from what clinicians wrote when we asked them to answer in ChatGPT's place. In Table~\ref{tab:guidelines}, we map these guidelines to the process failures they address and to a heuristic for each that can be used to determine whether a given response satisfies this guideline. 

\subsubsection{Guideline 1: Intentionally Sequence Responses to Psychological Distress}
Clinicians recommended sequencing responses to \textbf{focus on safety, co-regulation, then exploration, in that order}. We visualized this process in Figure ~\ref{fig:layers}. 
Risk language in any message should trigger a screening act before anything else follows. ChatGPT currently answers the first message of a distress conversation with content. In Nova's and Quinn's conversations it produced numbered steps within seconds of the disclosure and without asking anything. Risk language should be met with a careful and sensitive safety check, and what the system says next should depend on the answer that comes back. Sequencing may be difficult for current models, which ask clarifying questions three times less often than people do~\citep{Shaikh2025-gr} and commit early to solutions they then build on~\citep{Laban2025-dy}.

\subsubsection{Guideline 2: Hold Boundaries}
A GPCA should stay within its role, holding constant how it addresses the young person, its expressed closeness, and the role it claims to fill~\citep{Maeda2024-ga}.
ChatGPT currently modulates its responses based on the prompts it receives, and for example, 
in Sumaya's conversation it matched an intimate disclosure with an intimate response. 
Emotional responsiveness is valued by users and has improved markedly in recent models~\citep{Sabour2024-zu, Chen2023-nr, Sorin2024-ro}, but unconstrained responsiveness 
risks fostering dependency~\citep{namvarpour2026teenoverreliance, Zhang2025-ex} and can escalate the emotions of a young person already in distress. 
GPCAs should hold the boundary even as a user expresses greater intimacy or dysregulation.

\subsubsection{Guideline 3: Respond without Providing Advice}
A GPCA should be able to acknowledge a disclosure, ask a clarifying question, and stop without supplying unsolicited advice. ChatGPT instead defaults to offering extensive advice, lists, and topical completeness, even when a young person declines them, as Ace did twice. Almost all clinician rewrites contained no advice at all.

\subsubsection{Guideline 4: Do not make assumptions, and instead, state the limits of what the system knows.}
A GPCA should state what it does not know. ChatGPT currently makes \textit{assertive} claims about third parties it has not engaged with and claims about a young person's experience that the young person has not made. 
After stating these limitations, it should honor them and show humility in its response. In contrast, ChatGPT told Quinn it was not an expert but then supplied six recommendations. Prior work shows that
systems for young people must move from reading what someone says to recognizing why they are saying it~\citep{Blake2026-mf}, and marking what the system cannot know is a first step toward that. Clinicians' rewrites frequently asked users for more information and adopted a posture of curiosity rather than expertise.

\subsubsection{Guideline 5: Refer the User to Someone Reachable}
A referral should name a specific service or person the young person can actually reach. ChatGPT currently states referrals generically and treats it as an ending. Clinicians asked for the opposite on both counts: the resource should be presented directly instead of left for a young person in crisis to find, and the conversation should continue while they get there. Young people need several sources of support, and the immediacy and diversity of that network is itself protective ~\citep{Arnett2000-mb}.

Even if a GPCA follows all of these guidelines, a gap may remain between the support it can provide and the support that a human could provide in the same situation. Clinicians named some behaviors that do not translate easily into properties of a text-based system. They described reading nonverbal cues and co-regulating emotions with their clients. A GPCA can provide resources but it cannot reach out to a third party on a young person's behalf or supervise a successful handoff to them. Thus, we offer these guidelines not as a way to fully imitate or replace the support of a skilled human but as guidance for crafting safer responses and avoiding unintentional harm.

\subsection{Creating Guardrails for GPCAs}
Creating effective guardrails for GPCAs is an active topic of public discourse, policy discussion, and academic research. In the United States, many state legislatures are actively drafting policies to place design restrictions on CAs, such as disclaimers stating that the CA is not human or not an expert, nudges to limit use, or escalation to a human if a user expresses an intent to harm themselves (e.g., \cite{Padilla2026-zi, Curtis2026-cy}). Some of these policy efforts target minors specifically and would not apply to the interactions that we reviewed in our study. Other policy efforts target mental health support for users of any age~\cite{Barcott2026-ir}. For example, the state of Vermont recently passed a new bill preventing any AI system from offering mental health services or therapeutic communication without human oversight~\cite{Berbeco2026-rp}. These efforts are not specific to the United States. The European Union's AI Act, for example, requires, among other provisions, that the creators of the models that GPCAs build on conduct risk assessments of their systems to understand their potential to harm users seeking mental health support~\cite{European-CommissionUnknown-ix}.

Despite this widespread interest in requiring guardrails, designing them well is an active challenge. A meta-review examining research on AI for mental health identified ``safety and harm'' as a major theme across the scholarly literature and concludes that further work is needed to understand the risks that these systems pose, particularly in comparison to support from trained humans \cite{Rahsepar-Meadi2025-ha}. Other recent work synthesizing research on the governance of AI for mental health concludes that current CAs need additional oversight to avoid causing harm to people experiencing mental health challenges \cite{Lee2026-fn}.

Our work contributes to a growing evidence base that seeks to provide guidance in this context (e.g., \cite{yu2026safecompanions, Media2024-za, American-Psychological-Association2025-cv}). Our findings reveal that guardrails must go beyond flagging specific content as off limits. The clinicians in our study carefully calibrated their tone and emotional intensity and followed a predictable sequence within their responses to avoid causing harm. Many of these behaviors can be replicated by a text-based system. Each of the guidelines we propose deserves further validation in real-world contexts. Future work to validate a system that is designed to follow these guidelines (i.e., one that sticks to boundaries and a consistent persona, sequences responses as a clinician would, responds without providing advice, speaks with humility about its limitations, and provides concrete and accessible resources) would be a valuable next step. The guidelines we propose have potential to be integrated into GPCA regulations, provided they are first evaluated via in-the-wild studies.

\subsection{Limitations}

This study has at least two limitations. First, we only examined interactions that occurred on a single platform, ChatGPT. 
At the time of our study design, ChatGPT was the most used GPCA among young people. However, our interviews surfaced growing use of Google Gemini among this age group, and as with any technological domain, usage trends are likely to change over time. 
We do not know whether the patterns we report would hold for Gemini or any other system. The language models that underlie GPCAs are also updated frequently, which could change the characteristics of a system's responses.
Second, although we analyzed five conversations in depth and reviewed many others, these data cannot represent the full range of psychological distress that young people might discuss with a GPCA. Examining additional cases and collecting data from additional professionals would undoubtedly expand the set of insights we report here.

\section{Conclusion}
We studied how psychological distress relates to young people's use of ChatGPT across three analyses: a survey of 158 young adults and 19,930 conversations from
their chat histories, five conversations in which a young person brought distress
to ChatGPT, and a review of those conversations by ten clinicians who then rewrote them. 
We found distressed participants reported a closer relationship with ChatGPT and an
increased likelihood of following its advice than their peers. 
When they turned to ChatGPT in moments of acute distress, it was quick to give overly dramatic responses and excessive, action-oriented suggestions.
Clinicians endorsed ChatGPT's availability and much of its wording, and identified seven process failures, none concerning accuracy. 
Unlike ChatGPT's responses, clinicians' responses asked about safety, brought the intensity down, and only then explored. From these we derive five verifiable design guidelines for how GPCAs can respond to young people in
distress.

\bibliographystyle{ACM-Reference-Format}
\bibliography{aaai2026, paperpile}


\end{document}